\documentclass[runningheads]{llncs}
\usepackage{eccv}
\usepackage{eccvabbrv}
\usepackage{tikz}
\usepackage{multirow}
\usepackage{comment}
\usepackage{amsmath,amssymb}
\DeclareMathOperator*{\argmax}{argmax}

\usepackage{color}
\usepackage[misc]{ifsym}
\usepackage{url}
\usepackage{bm}
\usepackage{bbm}
\usepackage{soul}
\usepackage{cuted}
\usepackage{xcolor}
\usepackage{caption}
\usepackage{multirow}
\usepackage{mathrsfs}
\usepackage{algorithm}
\usepackage{algpseudocode}
\usepackage{graphicx}
\usepackage{booktabs}
\usepackage[accsupp]{axessibility} 

\usepackage{array}
\newcommand{\PreserveBackslash}[1]{\let\temp=\\#1\let\\=\temp}
\newcolumntype{C}[1]{>{\PreserveBackslash\centering}p{#1}}
\newcolumntype{R}[1]{>{\PreserveBackslash\raggedleft}p{#1}}
\newcolumntype{L}[1]{>{\PreserveBackslash\raggedright}p{#1}}

\usepackage[pagebackref,breaklinks,colorlinks]{hyperref}

\usepackage{orcidlink}

\begin{document}
\title{\textcolor{blue}{\textsf{DENOISER}}: Rethinking the Robustness \\ for Open-Vocabulary Action Recognition}

\titlerunning{Rethinking Robustness for OVAR}

\author{Haozhe Cheng\inst{1} \and
Cheng Ju\inst{1,2} \and
Haicheng Wang\inst{1} \and
Jinxiang Liu\inst{1} \and    \\
Mengting Chen\inst{2} \and
Qiang Hu\inst{1}  \and
Xiaoyun Zhang\inst{1}\,\textsuperscript{\Letter}  \and 
Yanfeng Wang\inst{1} 
}

\authorrunning{H. Cheng et al.}

\institute{$^1$Shanghai Jiao Tong University \quad \  \  $^2$Alibaba Group
\\  
\email{\{simoncheng,\,ju\_chen,\,xiaoyun.zhang\}@sjtu.edu.cn}
}

\maketitle 
\begin{abstract}
As one of the fundamental video tasks in computer vision, Open-Vocabulary Action Recognition (OVAR) recently gains increasing attention, with the development of vision-language pre-trainings. To enable generalization of arbitrary classes, existing methods treat class labels as text descriptions, then formulate OVAR as evaluating embedding similarity between visual samples and textual classes. However, one crucial issue is completely ignored: the class descriptions given by users may be noisy, {\em e.g.}, misspellings and typos, limiting the real-world practicality of vanilla OVAR. To fill the research gap, this paper pioneers to evaluate existing methods by simulating multi-level noises of various types, and reveals their poor robustness. To tackle the noisy OVAR task, we further propose one novel \textit{DENOISER} framework, covering two parts: generation and discrimination. Concretely, the generative part denoises noisy class-text names via one decoding process, {\em i.e.}, propose text candidates, then utilize inter-modal and intra-modal information to vote for the best. At the discriminative part, we use vanilla OVAR models to assign visual samples to class-text names, thus obtaining more semantics. For optimization, we alternately iterate between generative and discriminative parts for progressive refinements. The denoised text classes help OVAR models classify visual samples more accurately; in return, classified visual samples help better denoising. On three datasets, we carry out extensive experiments to show our superior robustness, and thorough ablations to dissect the effectiveness of each component.
\end{abstract}

\begin{figure}[t]
\begin{center}
\includegraphics[width=1\textwidth] {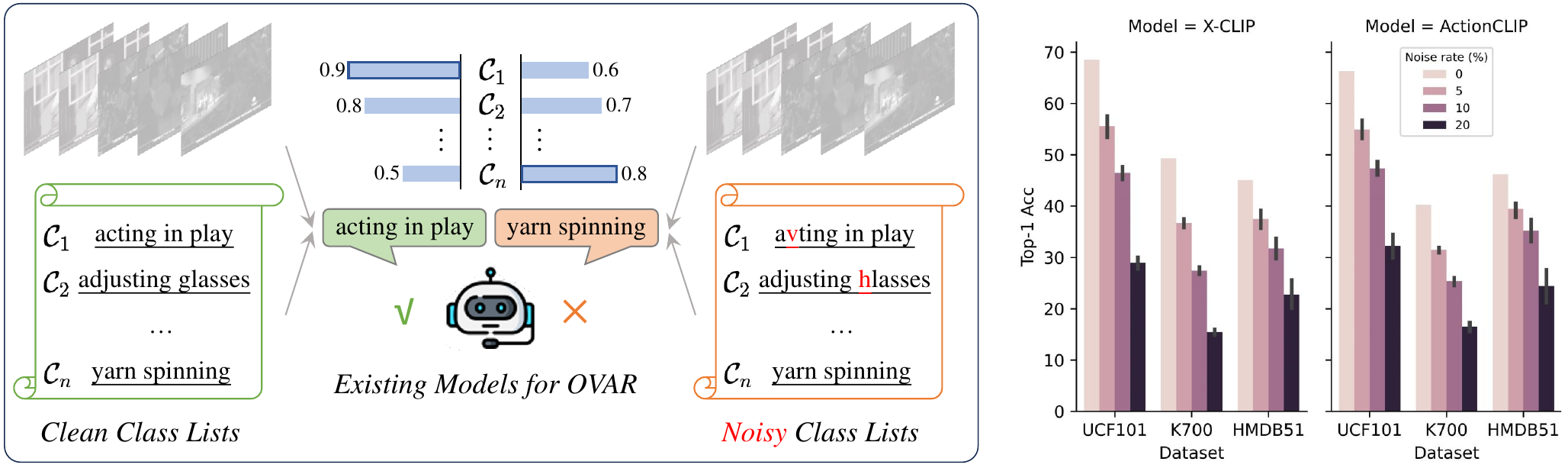}
\end{center}
\vspace{-0.5cm}
\caption{\textbf{Left}: For open-vocabulary action recognition (OVAR), existing researches neglect an essential aspect: the key class descriptions from users maybe noisy ({\em e.g.}, misspelling and typos), resulting in potential classification errors and limiting the real-world practicality. 
\textbf{Right:} Rethinking the robustness for popular methods~\cite{wang2021actionclip,zhou2023non}. On various datasets, these methods exhibit high sensitivity to noises. Besides, as the noise level increases, the performance degrades significantly.}
\vspace{-0.3cm}
\label{fig:intro}
\end{figure}

\vspace{-0.3cm}
\section{Introduction}  \label{sec:intro} 
Video action recognition is one of the fundamental tasks in computer vision that involves classifying one video into meaningful semantics. Despite huge progress that has been made, existing researches focus more on closed-set scenarios, where action categories remain constant during training and inference. Such scenarios are an oversimplification of real life, and thus limiting their practical application. Recently, another line of research considers one more challenging scenario, namely open-vocabulary action recognition (OVAR), and receives increasing attention. OVAR requires the model to handle one broader range of action categories, including novel (unseen) categories at inference time.

To tackle OVAR task, Vision-Language Alignment (VLA) paradigm~\cite{clip,Jia21,yuan2021florence} provides one preliminary but popular idea. By treating class names as textual descriptions, action recognition is formulated as one similarity between textual class embeddings and visual video embeddings. Following on this paradigm, recent works focus on minor improvements, {\em e.g.}, better align vision-language modalities~\cite{ju2022prompting,wang2021actionclip,zhou2023non}. Although promising, these researches all maintain one unrealistic assumption in real-world scenarios, {\em i.e.}, the given novel text categories are absolutely clean and accurate. The concrete form is that, they evaluate open-vocabulary performance by re-partitioning closed-set datasets in which class names are fully human-checked. But in fact, under OVAR, the novel class-list provided by users is similar to a text query. For thousands of queries, character misspellings (typos, missing, tense error) are inevitable~\cite{sakaguchi2017robsut,keller2021bert}, since users often don't double-check. Besides, differences in user habits and diversity of scenarios, can easily make text class descriptions somewhat noisy (Fig~\ref{fig:intro} Left).

We are hence motivated to fill the research gap of noisy class names in OVAR. For several standard datasets~\cite{Kuehne11,Soomro12,Carreira19}, we simulate various levels of noise, {\em i.e.}, randomly substituting, deleting and inserting words in novel class names with a certain probability. Fig.~\ref{fig:intro} Right empirically evaluates the noise effect for existing OVAR methods~\cite{ju2022prompting,wang2021actionclip,zhou2023non}. One can find that just a small amount of noise lowers classification accuracy by a large margin, implying quite poor robustness.

To spur the community to deal with the noisy OVAR task, being necessary and practical, this paper bravely faces the challenges. One vanilla idea is using a separate language model ({\em e.g.}, GPT~\cite{gpt3.5}) to correct noisy class descriptions, and then adapt the off-the-shelf vision-language paradigm~\cite{clip,Jia21,yuan2021florence}. However, there exist two nettlesome issues. \textbf{Textual Ambiguity}. Class name is usually a few compact words, with vague semantics, {\em e.g.}, for the noisy class text ``boird", there could be multiple cleaned candidates in terms of spelling, such as ``bird" and ``board". This short class name lacks context, making phrase correction difficult for uni-modal language models. \textbf{Cascaded Errors}. Text correction and action recognition are independently completed, without sharing knowledge. The noisy output of text correction is cascaded to the input of action recognition, resulting in continuous propagation of errors. To address these issues, we design one multi-modal robust framework: \textit{DENOISER}, for noisy OVAR.

Our first insight is to treat denoising of class texts as a \textit{generative} task: given noisy class texts, decode the clean ones. At \textit{generative} step, textual-visual information is taken into consideration to decide the best way of denoising. Specifically, it consists of three components: text proposals, inter-modal weighting and intra-modal weighting. We first propose potential candidates based on spelling similarity to limit the decoding space. Then, inter-modal weighting gives its vote according to the assigned visual samples; while intra-modal weighting uses solely textual information. Finally, the weights of both modules are combined to decide the best candidate. To achieve more semantics, at \textit{generative} step, we employ existing OVAR models to classify visual samples, then assign visual information to each noisy class text. In such way, visual samples help to clarify the semantic ambiguity of corresponding class texts, making full use of information.

To further avoid cascaded errors, we propose one solution of alternating iterations, to connect the \textit{generative} and \textit{discriminative} tasks. By progressive refinement, the denoised text classes help OVAR models to classify visual samples more accurately; the classified visual samples help better denoising. Under multiple iterations, the denoising results and OVAR are both better.

We carry out extensive experiments to show that our \textit{DENOISER} has superior robustness against noisy text labels in open-vocabulary action recognition. Our results can even approach that under clean class labels. Thorough ablation studies are carried out to dissect the effectiveness for each designs.

In general, our main contributions can be summarized as:

$\bullet$ We pioneer to explore noisy class texts for open-vocabulary action recognition (OVAR). By simulating noises of various types, we evaluate the robustness for existing work, and rethinking to study one practical task: noisy OVAR. 

$\bullet$ We propose a novel \textit{DENOISER} framework for noisy OVAR, to alternately optimize generative-discriminative parts. The generative part leverages decoding to denoises noisy class-text names, with visual-textual knowledge. The discriminative part assign classes to visual samples for action recognition.

$\bullet$ We extensively experiment to show the superior robustness of \textit{DENOISER} under various noises and datasets; full ablations are studied for components.

\section{Related Work} \label{sec:related}
{\noindent \bf Multi-Modal Pre-training} aims to jointly optimize vision-text-audio embeddings with large-scale web data, {\em e.g.}, CLIP~\cite{clip}, ALIGN~\cite{Jia21}, Florence~\cite{yuan2021florence}, FILIP~\cite{yao2021filip}, VideoCLIP~\cite{xu2021videoclip}, and LiT~\cite{zhai2022lit}. 
From a architecture perspective, the popular solution usually contains one visual encoder, one textual encoder and one audio encoder, followed by cross-modal fusion~\cite{liu2022exploiting,ju2023turbo}. 
From an optimization perspective, contrastive learning~\cite{chen2020learning,zheng2021contrastive} and cross-modal matching~\cite{cheng2023vindlu,li2022blip} are two mainstream branches, covering self supervision~\cite{liu2024annotation}, weak supervision~\cite{li2023blip,cheng2023mixer} and partial supervision~\cite{ju2023constraint,liu2024audio}. 
As a result, multi-modal pre-training benefits various potential applications: cross-modal retrieval~\cite{chen2023enhancing}, video understanding~\cite{ju2023distilling,ju2021divide,ju2020point,ju2022adaptive}, action recognition~\cite{ju2022prompting,zhao2020bottom}, visual grounding~\cite{liu2022exploiting,ye2021unsupervised,liu2023audio}, AIGC generation~\cite{ma2023diffusionseg,chen2024wear}.

\vspace{0.15cm}
{\noindent \bf Open-Vocabulary Concept Learning} aims to understand vision in terms of categories described by free textual descriptions. Relying on multi-modal pre-training, existing researches are mainly divided into three branches: recognition~\cite{wang2021actionclip,zhou2023non}, detection~\cite{ju2023multi,nag2022zero,yang2023multi} and segmentation~\cite{wang2023towards,ma2024open,ma2023attrseg}. However, they assume textual descriptions to be absolutely clean as it is a common practice to evaluate open-vocabulary performance by re-partitioning closed-set datasets whose class names are human-checked, limiting real-world application. This paper pioneers to take noises from text labels (misspellings and typos) into consideration, and propose one \textit{DENOISER} framework for solution.

\vspace{0.15cm}
{\noindent \bf Robustness of Language Models} is extensively studied in the field of adversarial attacks and their defense techniques~\cite{wang2019towards,zhang2020adversarial}. Facing adversarial perturbation in text inputs, the defense methods help the model output correctly as if the inputs are not perturbed. They can be categorized into two groups: detection and purification~\cite{zhou2019learning, pruthi2019combating}, as well as adversarial training~\cite{xu2019lexicalat,dinan2019build,liu2020robust, liu2020joint,wang2020defense}. The former ones use sophisticated networks to model the context so as to detect the corrupted part of a phrase and correct it. The latter one intends to train the model on adversarial samples so that they could learn how to output the correct answers, when faced with adversarial text inputs. Overall, they employ solely textual information to robustify the model. We differ from them by leveraging multi-modal information to better assist the text denoising.

\vspace{0.2cm}
{\noindent \bf Multi-Modal Robust Learning} aims to help the vision-language foundation models learn from noisy correspondence at training time, so as to mitigate the well-known problem that scrapped text-images pairs from the internet suffers from mismatched correspondence~\cite{hu2021learning}. Our work differs from them by considering the possible noisy text labels at testing time.

\begin{figure}[t]
\centering
\vspace{0.2cm}
\includegraphics[width=\linewidth]{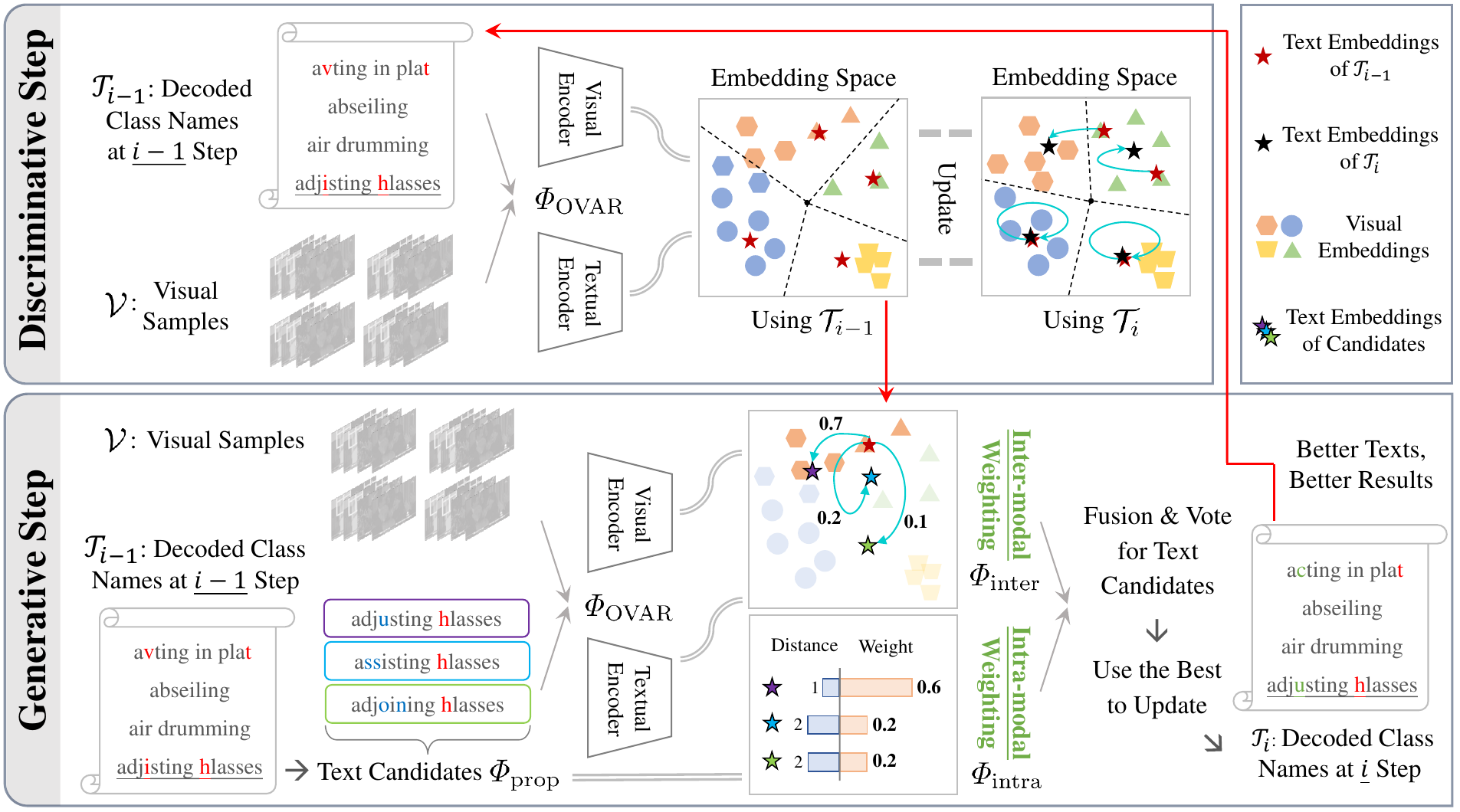}
\vspace{-0.5cm}
\caption{\textbf{Framework Overview}. \textit{DENOISER} is composed of one \textit{generative} part $\Psi_{\mathrm{gene}}$ and one \textit{discriminative} part $\Psi_{\mathrm{disc}}$. $\Psi_{\mathrm{gene}}$ views denoising the text labels as a decoding process $\mathcal{T}_{i-1}\rightarrow\mathcal{T}_{i}$. We first propose text candidates $\Phi_{\mathrm{prop}}$ for $\mathcal{T}_{i-1}$ based on spelling similarity; then choose the best candidate by inter-modal weighting $\Phi_{\mathrm{inter}}$ and intra-modal weighting $\Phi_{\mathrm{intra}}$. $\Phi_{\mathrm{inter}}$ uses visual-textual information, while $\Phi_{\mathrm{intra}}$ relies solely on texts to vote. $\Psi_{\mathrm{disc}}$ assigns classes to visual samples. Then only visual samples that match classes can vote for text candidates, making better usage of classes. We optimize alternatively between \textit{generative} and \textit{discriminative} steps to tackle noisy OVAR.}
\vspace{-0.2cm}
\label{fig:Framework}
\end{figure}

\section{Method} \label{sec:method} 
This paper explores the noisy text classes for open-vocabulary action recognition. Concretely, in Sec~\ref{preliminary}, we introduce the noisy open-vocabulary setting; in Sec~\ref{MRC}, we detail our \textit{DENOISER} framework, containing \textit{generative} - \textit{discriminative} sub-tasks; in Sec~\ref{em}, we report the accompanying optimization strategy.

\subsection{Preliminary \& Rethinking}  \label{preliminary}
\noindent \textbf{Open-Vocabulary Action Recognition (OVAR).} 
Given one video dataset $\mathcal{V} = (v_j\in \mathbb{R}^{T \times H\times W\times 3})_j^{N}$, OVAR aims to train one model $\Phi_{\mathrm{OVAR}}$ that concludes target videos into arbitrary classes (in the form of text description $\mathcal{T}$). 
\begin{equation}
  \mathcal{Y}^{\mathrm{train}} = \Phi_{\mathrm{OVAR}}(\mathcal{V}^{\mathrm{train}}\, ,\mathcal{T}) \in \mathbb{R}^{C_{\text{base}}}, \quad
  \mathcal{Y}^{\mathrm{test}} = \Phi_{\mathrm{OVAR}}(\mathcal{V}^{\mathrm{test}}\, , \mathcal{T}) \in \mathbb{R}^{C_{\text{novel}}}.  
\end{equation}
During training, (video, class label) pairs from the base class are provided, {\em i.e.}, $\{(\mathcal{V}^{\text{train}}, \mathcal{Y}^{\text{train}}) \sim C_{\text{base}}\}$; while during testing, the model is evaluated on the disjoint novel classes, {\em i.e.}, $\{(\mathcal{V}^{\text{test}}, \mathcal{Y}^{\text{test}}) \sim C_{\text{novel}}\}$. Note that, training classes $C_{\text{base}}$ and testing class $C_{\text{novel}}$ are disjoint, \ie, $C_{\text{base}} \cap C_{\text{novel}} = \varnothing$.

\vspace{0.2cm}
\noindent \textbf{Vision-Language Alignment (VLA).} \ 
To enable open-vocabulary capability, recent OVAR studies~\cite{ju2022prompting,wang2021actionclip,zhou2023non,qian2022multimodal} embrace vision-language pre-trainings (VLPs), for their notable ability in cross-modal alignment. Specifically, regarding vanilla class names $C \ (C_{\text{base}}/C_{\text{novel}})$ as the textual descriptions $\mathcal{T} \ (\mathcal{T}_{\text{base}}/\mathcal{T}_{\text{novel}})$, OVAR could be achieved by measuring the embedding similarity between class-level text and video-level vision. Formally, these methods are formulated as: 
\begin{equation}
    \mathcal{Y} = \sigma(\mathcal{F}_v * \mathcal{F}_t), \quad
    \mathcal{F}_v = \Phi_{\mathrm{pool}}(\Phi_{\mathrm{vis}}(\mathcal{V})) \in\mathbb{R}^{N \times D}, \quad
    \mathcal{F}_t = \Phi_{\mathrm{txt}}(\mathcal{T}) \in\mathbb{R}^{C \times D}.
\end{equation}
where $\sigma$ refers to softmax activation, $\Phi_{\mathrm{pool}}$ is spatio-temporal pooling, $\Phi_{\mathrm{vis}}$ and $\Phi_{\mathrm{txt}}$ are visual and textual encoders of VLPs, $D$ is the embedding dimension.

\vspace{0.2cm}
\noindent \textbf{Noisy Text Descriptions in OVAR.} 
Although great progress has been made, the VLA paradigm suffers from an unrealistic assumption, {\em i.e.}, text descriptions describing classes are absolutely clean/accurate, limiting the practicality in reality. Actually, the diversity of users and scenarios can easily make text descriptions of novel classes somewhat noisy. Formally, for class $c$, assuming its text description has $n$ words, the clean version $\mathcal{T}_c$ and noisy version $\mathcal{T}_c^\prime$ are:
\begin{equation}   \label{eq:noise}
\mathcal{T}_c^\prime=(t_{c,1}^\prime,\cdots,t_{c,n}^\prime) = \Psi_{\mathrm{noise}}(\mathcal{T}_c\, ; p),  \quad  \mathcal{T}_c=(t_{c,1},\cdots,t_{c,n}).
\end{equation} 
$\Psi_{\mathrm{noise}}$ refers to the process of noise contamination in reality, {\em e.g.}, {\em inserting}, {\em substituting} and {\em deleting} letters with probability $p$, following \cite{rychalska2019models,sun2023benchmarking}. Here, we consider character-level noise for its generality, let $\mathcal{T}_c^\prime$ has the same number of words as $\mathcal{T}_c$ for simplicity. Note that, in the following manuscript, we reasonably ignore the subscript $c$, {\em e.g.}, denoting $\mathcal{T}^\prime = \{\mathcal{T}_c^\prime\}_c$ and $t_i = \{t_{c,i}\}_c$, when class is arbitrary.

As a result, the noisy OVAR task can be formulated as: given $\mathcal{V}$ and $\mathcal{T}^\prime$, the model is expected to maximize classification accuracy, and recovering $\mathcal{T}$.

\vspace{0.2cm}
\noindent \textbf{Robustness of Existing Methods.} \ 
Fig.~\ref{fig:intro} evaluates for OVAR studies~\cite{wang2021actionclip,zhou2023non}, across three public datasets. In terms of Top-1 classification accuracy, previous methods are rather sensitive to noise. It shows one trend: the larger the noise, the more significant the performance degradation. 
Such poor robustness proves excessive idealization of existing work, also motivates us to fill the research gap.

\subsection{DENOISER: Robust OVAR Framework}  \label{MRC}
Here, we propose one robust framework for noisy OVAR, namely \underline{DENOISER}. 

\vspace{0.2cm}
\noindent \textbf{Motivation.} \ 
Given the complexity of noisy OVAR, we here divide it into two sub-tasks: denoising of class texts, and vanilla OVAR. 
The former is viewed as a \underline{\textit{generative}} decoding form, by taking both vision and text into consideration for good denoising. While the latter is in a natural \underline{\textit{discriminative}} form, by assigning text class labels to visual samples. 
For the joint optimization of these two sub-tasks, we iterate alternately between \textit{generative} and \textit{discriminative} forms. As a result, our \textit{DENOISER} framework tackles noisy OVAR progressively.

\vspace{0.2cm}
\noindent \textbf{Framework.}
As shown in Fig.~\ref{fig:Framework}, our \textit{DENOISER} framework covers two components: \textit{generative}  sub-task $\Psi_{\mathrm{gene}}$ and \textit{discriminative}  sub-task $\Psi_{\mathrm{disc}}$.

For $\Psi_{\mathrm{gene}}$, we iteratively refine the class texts by one decoding process, that is, $(\mathcal{T}_0, \mathcal{T}_1, \cdots, \mathcal{T}_n)$, where $n$ is the index of decoding steps. Upon finishing step $i$, we will have $\mathcal{T}_{i} = (\overline{t_{1}},\cdots,\overline{t_{i}},t_{i+1}^\prime,\cdots,t_{n}^\prime)$, where $\overline{t}$ refers to the decoded version of $t$, meaning that the $i$-th word of text labels is decoded at step $i$. We start with $\mathcal{T}_0 = \mathcal{T}^\prime$, and finish at $\mathcal{T}_n$ to ensure that all words are denoised. While for $\Psi_{\mathrm{disc}}$, we find it identical to vanilla OVAR task, and thus leveraging the VLA pipeline~\cite{ju2022prompting,wang2021actionclip} for help, which is off-the-shelf and well-studied. Formally, our \textit{DENOISER} framework tackles noisy OVAR as follows: 
\begin{align}
    \mathcal{T}_{i} = \Psi_{\mathrm{gene}}(\mathcal{T}_{i-1}, \mathcal{Y}, \mathcal{V}),  \quad 
    \mathcal{Y} = \Psi_{\mathrm{disc}}(\mathcal{T}_{i-1}, \mathcal{V}) = \Phi_{\mathrm{OVAR}}(\mathcal{T}_{i-1}, \mathcal{V}).
\end{align}

At \textit{discriminative} part, we assign class labels $\mathcal{Y}$ to $\mathcal{V}$, which is then used in the \textit{generative} step. At the \textit{generative} part, we first propose $K$ text candidates $\Phi_{\mathrm{prop}}(\mathcal{T}_{i-1})$ for $\mathcal{T}_{i}$ base on $\mathcal{T}_{i-1}$ to limit the decoding space. Then, to decide the best candidate, we design two novel solutions, namely inter-modal weighting $\Phi_{\mathrm{inter}}$ and intra-modal weighting $\Phi_{\mathrm{intra}}$. Here, $\Phi_{\mathrm{inter}}$ employs visual information $\mathcal{V}$ and assigned labels $\mathcal{Y}$, while $\Phi_{\mathrm{intra}}$ relies solely on textual information $\mathcal{T}_{i-1}$.

We alternate between the \textit{generative} and \textit{discriminative} steps to optimize the decoding result step by step. Please find in Algorithm \ref{algo} for details.

\subsection{Optimization for \textit{DENOISER} Framework}  \label{em} 
\noindent \textbf{Discriminative Step} consists in assigning class labels $\mathcal{Y}$ using $\Psi_{\mathrm{disc}}$. Intuitively, visual samples $\mathcal{V}_c$ whose labels $\mathcal{Y}$ are assigned to class $c$, {\em i.e.}~$\argmax{\mathcal{Y}}=c$, are those who could help decode $\mathcal{T}_{c,i}$ most efficiently. On the contrary, visual samples from other categories may have few connections with the current category $c$ and thus provide no meaningful aid. Here, we find this process is identical to vanilla OVAR, and hence employs $\Phi_{\mathrm{OVAR}}$ as $\Psi_{\mathrm{disc}}$. We theoretically prove in Appendix that, $\mathcal{V}_c$ is the best set of visual samples to choose from.

With $\mathcal{V}_c$ defined and $\argmax{\mathcal{Y}}=c$, $\Psi_{\mathrm{gene}}$ decodes $\mathcal{T}_{c,i}$ for each class $c$: 
\begin{align}
\Psi_{\mathrm{gene}}(\mathcal{T}_{c,i-1}, \mathcal{Y}, \mathcal{V}) = \Psi_{\mathrm{gene}}(\mathcal{T}_{c,i-1}, \mathcal{V}_c) = \argmax_{\mathcal{T}_{c,i}} p(\mathcal{T}_{c,i}|\mathcal{T}_{c,i-1},\mathcal{V}_c).
\end{align}
Recall that $t_{c, i}$ is the $i$-th word to be decoded of class text of category $c$, and that $\mathcal{T}_{c, i-1}$ is the class text of category $c$ from last decoding step, with the first $i-1$ words decoded. As we decode word-by-word, choosing the best $\mathcal{T}_{c,i}$ is exactly choosing the best $t_{c,i}$, {\em i.e.}~$\argmax_{\mathcal{T}_{c,i}} p(\mathcal{T}_{c,i}|\mathcal{T}_{c,i-1},\mathcal{V}_c) = \argmax_{t_{c,i}} p(t_{c,i}|\mathcal{T}_{c,i-1},\mathcal{V}_c)$ which we do in the \textit{generative} step.

\vspace{0.2cm}
\noindent \textbf{Generative Step} consists in, for each category $c$, choosing the best $t_{c,i}$ that maximizes $p(t_{c,i} | \mathcal{T}_{c, i-1}, \mathcal{V}_c)$. With $p(\mathcal{T}_{c, i-1}, \mathcal{V}_c)$ and $p(\mathcal{V}_c)$ same for all possible $t_{c,i}$, we make detailed derivations in Appendix to show that: 
\begin{align}
p(t_{c,i} | \mathcal{T}_{c,i-1}, \mathcal{V}_c) \varpropto p(t_{c,i},  \mathcal{T}_{c,i-1}, \mathcal{V}_c) \varpropto \prod_{v_j\in \mathcal{V}_c} p(t_{c,i}|v_j) p(\mathcal{T}_{c, i-1}|t_{c,i}, v_j).
\end{align}
Here, the error model $p(\mathcal{T}_{c, i-1}|t_{c,i}, v_j) $ evaluates how $t_{c,i}$ may be misspelled as $t_{c,i}^\prime$, since the $i$-th word in $\mathcal{T}_{c, i-1}$ is still noisy and not decoded. Knowing that errors made in text descriptions are independent of visual samples, it reduces to uni-modal $p(\mathcal{T}_{c, i-1}|t_{c,i})$. As the error that one may make given the correct text is harder to model while the reverse is much easier, we let $p(\mathcal{T}_{c, i-1}|t_{c,i}) \varpropto  p(t_{c,i}|\mathcal{T}_{c, i-1})$. Please refer to detailed derivations in Appendix. As a result, our final objective can be written as: 
\begin{align}
p(t_{c,i}|\mathcal{T}_{c, i-1}) \prod_{v_j\in \mathcal{V}_c} p(t_{c,i}|v_j) = \Phi_{\mathrm{intra}} \prod_{v_j\in \mathcal{V}_c} \Phi_{\mathrm{inter}} .
\end{align}

\vspace{0.15cm}
\noindent \underline{\textit{Text Proposals}} consists in proposing $K$ candidates $\{t_i^k\}_k$ for $t_i$ with the lowest Levenshtein Edit Distance $\mathcal{D}(\cdot, t_i^\prime)$, a metric of similarity in terms of spelling with the noisy word $t_i^\prime$. By replacing original noisy word $t_i^\prime$ in $\mathcal{T}^k_{i-1}$ with $\{t_i^k\}_k$, they form $\Phi_{\mathrm{prop}}(\mathcal{T}_{i-1}) = \mathcal{T}^k_{i} = (\overline{t_1},\cdots,\overline{t_{i-1}},t^k_i,t_{i+1}^\prime,\cdots,t_n^\prime)$, the $K$ candidates for $\mathcal{T}_{i}$. The benefit of text proposal is to reduce computation complexity. Since text embeddings are quantized in the semantic space, there is no need to search in the whole space. Instead, we limit to the proposed candidates.

\begin{algorithm}[t]
\caption{Denoiser: Robust Open-Vocabulary Action Recognition}  \label{alg:cap}
\begin{algorithmic}
\Require noisy text descriptions $\mathcal{T}^\prime$, visual samples $\mathcal{V}$, iteration number $n$, temperature $\lambda$, candidate number $K$, edit distance $\mathcal{D}$, open-vocabulary model $\Phi_{\mathrm{OVAR}}$ 
\State $\mathcal{T}_0 \gets \mathcal{T}^\prime$
\For {$i=1,2,\cdots,n$}
    \For {$c=1,2,\cdots,C$} \Comment{Text proposals}
        \State $t_{c,i}^\prime$ is the $i$-th word of $\mathcal{T}_{c,i-1}$, which is noisy and not yet decoded
        \State Select from corpus, $K$ candidates $\{t_{c,i}^k\}_k$ with the smallest $\mathcal{D}$ with $t_{c,i}^\prime$
        \State Replace $t_{c,i}^\prime$ with $\{t_{c,i}^k\}_k$, forming $\{\mathcal{T}_{c,i}^k\}_k$
    \EndFor
    \For {$j=1,2,\cdots,|\mathcal{V}|$} \Comment{Discriminative Step}
        \State $c \gets \argmax\limits_c \max\limits_k \frac{\text{exp}(\mathcal{S}(v_j,\mathcal{T}^k_{c, i}))}{\sum_{k^\prime} \text{exp}(\mathcal{S}(v_j,\mathcal{T}^{k^\prime}_{c, i}))}$
        \State Assign $v_j$ to class $c$, $v_j \in \mathcal{V}_{c}$
    \EndFor
    \For {$c=1,2,\cdots,C$} \Comment{Generative Step}
        \State $\Phi_{\mathrm{intra}}^k \gets \frac{\text{exp} (-\mathcal{D}(t^k_{c,i}, t^\prime_{c,i})/\lambda)}{\sum_{k^\prime} \text{exp} (-\mathcal{D}(t^{k^\prime}_{c,i}, t^\prime_{c,i})/\lambda)}$ \Comment{Intra-Modal Weighting}
        \State $\Phi_{\mathrm{inter}}^k \gets \prod_{v_j \in \mathcal{V}_c}\frac{ \text{exp}(\mathcal{S}(v_j,\mathcal{T}^k_{c, i}))}{\sum_{k^\prime} \text{exp}(\mathcal{S}(v_j,\mathcal{T}^{k^\prime}_{c, i}))}$ \Comment{Inter-Modal Weighting}
        \State $\mathcal{T}_{c,i} \gets \mathcal{T}_{c,i}^k$, $k = \argmax_k \Phi_{\mathrm{intra}}^k \times \Phi_{\mathrm{inter}}^k$
    \EndFor
\EndFor
\end{algorithmic}
\label{algo}
\end{algorithm}

\vspace{0.15cm}
\noindent \underline{\textit{Inter-modal Weighting}} 
$\Phi_{\mathrm{inter}}=p(t_{c,i}|v_j), ~ v_j \in \mathcal{V}_{c}$ relies on visual information assigned to class $c$ to determine the best $t_{c,i}$ for the next iteration. Concretely, we model the probability of being chosen for each proposed candidate to be:
\begin{align}
\mathbb{P}(t_{c,i} = t^k_{c,i}|v_j) = \mathbb{P}(\mathcal{T}_{c,i} = \mathcal{T}^k_{c,i}|v_j) = \frac{ \text{exp}(\mathcal{S}(v_j,\mathcal{T}^k_{c, i}))}{\sum_{k^\prime} \text{exp}(\mathcal{S}(v_j,\mathcal{T}^{k^\prime}_{c, i}))}, ~ v_j \in \mathcal{V}_{c}.
\end{align}
where $\mathcal{S}(\cdot, \cdot)$ represents the cosine similarity between embedding of a visual sample and a text, both encoded by $\Phi_{\mathrm{OVAR}}$. The intuition is that the more unanimously visual samples agree on a certain candidate $\mathcal{T}^{k}_{c, i}$, the more likely it is the corresponding class text of that category $c$. Furthermore, by making visual samples to vote on $\mathcal{T}_{c,i}^k$ instead of $t_{c,i}^k$, we take into consideration not only the current word $t_{c,i}$ but also the context implicitly.

\vspace{0.15cm}
\noindent \underline{\textit{Intra-modal Weighting}} $\Phi_{\mathrm{intra}} = p(t_{c,i}|\mathcal{T}_{c, i-1})$ relies solely on the textual information to decide the best $t_{c,i}$ for next iteration. To achieve the goal without introducing complex language models, we design a simple model by considering only spelling similarity and ignoring contexts, {\em i.e.}, by letting the choice of $t_{c,i}$ to depend solely on $t^\prime_{c,i}$ instead of on the entire $\mathcal{T}_{c,i-1}$: 
\begin{align}
\mathbb{P}(t_{c,i} = t_{c,i}^k|\mathcal{T}_{c,i-1}) = \mathbb{P}(t_{c,i} = t_{c,i}^k|t^\prime_{c,i}) = \frac{\text{exp} (-\mathcal{D}(t^k_{c,i}, t^\prime_{c,i})/\lambda)}{\sum_{k^\prime} \text{exp} (-\mathcal{D}(t^{k^\prime}_{c,i}, t^\prime_{c,i})/\lambda)}.
\end{align}
The intuition is that, the more similar a word candidate $t^k_{c,i}$ is, compared to the noisy word $t^{\prime}_{c,i}$, the more likely it is the corresponding denoised word. Here, we introduce one temperature parameter $\lambda$ to balance $\Phi_{\mathrm{intra}}$ and $\Phi_{\mathrm{inter}}$. A larger $\lambda$ indicates that different edit distance gives similar probabilities, meaning that we rely more on visual samples for decision, and vice versa.

\section{Experiments}  \label{sec:experiments}
\subsection{Setup}  \label{sec:setup}
\textbf{Typical Models for Vanilla OVAR}. 
To illustrate the generalizability of our framework, we leverage two typical models from the VLA pipeline as $\Phi_{\mathrm{OVAR}}$, that is, \underline{ActionCLIP}~\cite{wang2021actionclip} and \underline{XCLIP}~\cite{zhou2023non}. These two models adopt hand-crafted prompts and visual-conditioned prompt tuning, respectively. Under both models, we choose ViT-B/16-32F as the network backbones, for simplicity.

\begin{table}[t]
\centering
\caption{\textbf{Comparisons to Competitors.} On UCF101 and using ActionCLIP~\cite{wang2021actionclip} as $\Phi_{\mathrm{OVAR}}$, we compare with statistical text spell-checkers (PySpellChecker~\cite{pyspellchecker}), neural based ones (Bert of NeuSpell)~\cite{Jayanthi2020NeuSpellAN}, and GPT\,3.5~\cite{gpt3.5}. Our method remarkably outperforms others in terms of classification accuracy, semantic similarity of recovered class text, and performance of $\Phi_{\mathrm{OVAR}}$ under the noisy OVAR setting.}
\vspace{-0.2cm}
\begin{tabular}{c|C{3.1cm}|C{1.7cm}C{1.7cm}C{3cm}}
\toprule
\begin{tabular}[c]{@{}c@{}}Noise Rate (\%)\end{tabular} & Competitor & Top-1 Acc  & Label Acc  & Semantic Similarity \\ \midrule \midrule
0                     & Upper Bound  & 66.3 & 100 & 100\  \\ \midrule
\multirow{4}{*}{5} & GPT\,3.5~\cite{gpt3.5}    & $59.7_{\pm1.2}$ &  $47.6_{\pm3.1}\ $   &  $95.9_{\pm0.4}\ $   \\
                      & Bert (NeuSpell)~\cite{Jayanthi2020NeuSpellAN}         & $56.6_{\pm0.5}\ $ &  $66.2_{\pm2.3}\ $   &  $94.6_{\pm0.4}\ $   \\
                      & PySpellChecker~\cite{pyspellchecker} & $60.9_{\pm1.1}\ $ &  $82.5_{\pm2.9}\ $   &  $97.1_{\pm0.4}\ $   \\ 
                      & \textbf{Ours} & \boldsymbol{$63.8_{\pm0.7}\ $} &  \boldsymbol{$86.4_{\pm2.3}\ $}   &  \boldsymbol{$97.7_{\pm0.2}\ $}  \\
\midrule
\multirow{4}{*}{10}  & GPT\,3.5~\cite{gpt3.5}        & $58.5_{\pm1.3}\ $ &  $51.6_{\pm2.3}\ $   &  $95.8_{\pm0.3}\ $  \\
                      & Bert (NeuSpell)~\cite{Jayanthi2020NeuSpellAN}         & $51.0_{\pm0.5}\ $ &  $50.4_{\pm3.6}\ $   &  $91.6_{\pm0.6}\ $  \\
                      & PySpellChecker~\cite{pyspellchecker} & $55.7_{\pm1.1}\ $ &  $69.3_{\pm1.5}\ $   &  $94.8_{\pm0.3}\ $  \\
                      & \textbf{Ours} & \boldsymbol{$61.2_{\pm0.8}\ $} &  \boldsymbol{$75.9_{\pm1.9}\ $}   &  \boldsymbol{$96.4_{\pm0.3}\ $}  \\
\midrule
\multirow{4}{*}{20} & GPT\,3.5~\cite{gpt3.5}        & \boldsymbol{$57.5_{\pm1.8}\ $} &  $53.0_{\pm3.5}\ $   &  \boldsymbol{$95.1_{\pm0.6}\ $}  \\
                      & Bert (NeuSpell)~\cite{Jayanthi2020NeuSpellAN}         & $38.1_{\pm0.7}\ $ &  $26.6_{\pm2.7}\ $   &  $85.9_{\pm0.4}\ $  \\
                      & PySpellChecker~\cite{pyspellchecker} & $48.9_{\pm2.3}\ $ &  $46.5_{\pm3.1}\ $   &  $90.9_{\pm0.6}\ $  \\
                      & \textbf{Ours} & $54.2_{\pm1.3}\ $ &  \boldsymbol{$53.7_{\pm2.6}\ $}   &  $92.4_{\pm0.5}\ $  \\
\bottomrule
\end{tabular}
\label{table:baselines}
\end{table}

\vspace{0.1cm}
\noindent \textbf{Datasets}. 
\underline{HMDB-51}~\cite{Kuehne11} contains 7k videos of action $51$ categories. \underline{UCF-101}~\cite{Soomro12} covers 13k videos spanning $101$ categories. \underline{Kinetics-700}~\cite{Carreira19} (K-700) is simply an extension of K-400, with around 650k video clips sourced from YouTube. To partition these datasets for open-vocabulary action recognition, this paper follows the standard consensus~\cite{wang2021actionclip,zhou2023non}, for the sake of fairness.

\vspace{0.1cm}
\noindent \textbf{Noise.} \ As shown in Eq.~(\ref{eq:noise}), we add noise to class labels with rate $p$. Specifically, for each character of one text description, there is a probability $p$ that we noise it, {\em i.e.}~substitute it, delete it or insert behind it a new character.

\vspace{0.1cm}
\noindent \textbf{Metric.} 
We here propose three metrics from multiple perspectives, to comprehensively evaluate our framework. \underline{Top-1 Acc} refers to the top-1 classification accuracy, showing the robustness that our framework adds to the vanilla OVAR methods. \underline{Label Acc} calculates the percentage of denoised class-text labels that match exactly with the cleaned ground truth. \underline{Semantic Similarity} measures the cosine similarity of text embeddings, between denoised text labels and ground-truth text labels. Note that, Label Acc and Semantic Similarity evaluate directly how well the noisy text-class labels are recovered.

\begin{table}[t]
\centering
\caption{\textbf{Comparison Across Datasets and Models}. On three standard datasets, under various noise rates, our \textit{DENOISER} consistently improves the performance of $\Phi_{\mathrm{OVAR}}$ in noisy OVAR settings regardless of underlying OVAR methods $\Phi_{\mathrm{OVAR}}$.}
\vspace{-0.2cm}
\begin{tabular}{c|C{1cm}|C{2cm}C{2cm}C{2cm}C{2cm}}
\toprule
\multirow{3}{*}{Dataset} & \multirow{3}{*}{\shortstack[c]{Noise \\ Rate \\ \%}} & \multicolumn{4}{c}{$\Phi_{\mathrm{OVAR}}$: Typical Models for Vanilla OVAR task} \\ \cmidrule(r){3-6} 
 &  & \multicolumn{2}{c|}{ActionCLIP~\cite{wang2021actionclip}} & \multicolumn{2}{c}{XCLIP\cite{zhou2023non}} \\ \cmidrule(r){3-6} 
 &  & w/o Ours & \multicolumn{1}{c|}{\textbf{w Ours}} & w/o Ours  & \textbf{w Ours} \\ \hline \hline
\multirow{4}{*}{UCF101} & 0 & \boldsymbol{$66.3$} & \multicolumn{1}{C{2cm}|}{$65.1$} & \boldsymbol{$68.6$} & $66.9$ \\
 & 5 & $54.9_{\pm1.8}$  & \multicolumn{1}{c|}{\boldsymbol{$63.2_{\pm0.7}$}} & $55.6_{\pm2.2}$ & \boldsymbol{$64.2_{\pm1.4}$} \\
 & 10 & $47.3_{\pm1.4}$ & \multicolumn{1}{c|}{\boldsymbol{$61.2_{\pm1.2}$}} & $46.4_{\pm1.3}$  & \boldsymbol{$62.9_{\pm2.3}$} \\
 & 20 & $32.3_{\pm2.4}$ & \multicolumn{1}{c|}{\boldsymbol{$54.8_{\pm2.6}$}} & $29.0_{\pm1.2}$  &  \boldsymbol{$55.9_{\pm2.6}$}\\
 \midrule
\multirow{4}{*}{HMDB51} & 0 & \textbf{46.2} & \multicolumn{1}{c|}{$42.8$} & \textbf{45.0} & $42.0$ \\
 & 5 & $39.4_{\pm1.4}$  & \multicolumn{1}{c|}{\boldsymbol{$41.3_{\pm1.4}$}} & $37.5_{\pm1.8}$  & \boldsymbol{$39.7_{\pm1.0}$} \\
 & 10 & $35.2_{\pm2.3}$ & \multicolumn{1}{c|}{\boldsymbol{$39.6_{\pm1.4}$}} & $31.8_{\pm2.2}$ & \boldsymbol{$37.3_{\pm1.5}$} \\
 & 20 & $24.4_{\pm3.3}$ & \multicolumn{1}{c|}{\boldsymbol{$35.5_{\pm3.2}$}} & $22.7_{\pm2.8}$  & \boldsymbol{$30.9_{\pm1.6}$} \\
 \midrule
\multirow{4}{*}{K700} & 0 & \textbf{40.2}  & \multicolumn{1}{c|}{$38.7$} & \textbf{49.3}  & $47.5$ \\
 & 5 & $31.5_{\pm0.5}$ & \multicolumn{1}{c|}{\boldsymbol{$36.8_{\pm0.3}$}} & $36.7_{\pm0.9}$  & \boldsymbol{$44.1_{\pm0.6}$} \\
 & 10 & $25.4_{\pm0.8}$ & \multicolumn{1}{c|}{\boldsymbol{$35.3_{\pm0.5}$}} & $27.5_{\pm0.7}$  & \boldsymbol{$41.8_{\pm0.9}$} \\
 & 20 & $16.5_{\pm0.9}$ & \multicolumn{1}{c|}{\boldsymbol{$29.5_{\pm0.8}$}} & $15.5_{\pm0.6}$  & \boldsymbol{$34.6_{\pm0.4}$} \\ \bottomrule[0.3mm]
\end{tabular}
\label{table:main}
\end{table}

\vspace{0.1cm}
\noindent \textbf{Implementation Details.} 
We here set the number of text candidates $K=10$. Intra-modal weighting and inter-modal weighting are both used to determine the best candidate. Temperature $\lambda$ follows a linear schedule from 0.01 to 1. We use the same corpus as in PySpellChecker, which contains 70317 English words, for text proposals. For typical OVAR methods~\cite{wang2021actionclip,zhou2023non}, we choose the ViT-B/16-32F checkpoint pretrained on K400~\cite{K400} to evaluate their zero-shot robustness on HMDB51~\cite{HMDB}, UCF101~\cite{UCF101} and K700~\cite{K700}. Since K700 and K400 have overlapped categories, we exclude them when evaluating on K700. For UCF101, we use the separated lowercase text label. All ablation studies are conducted on UCF101 under 20\% noise. In pursuit of statistical significance, We do each simulation 10 times and report the mean and confidence interval of 95\%.

\subsection{Comparison with State-of-the-art Methods}  \label{sec:comparison}
\noindent \textbf{Comparison to Competitors.} 
Table \ref{table:baselines} compares different baselines from three axes: Top-1 Acc of $\Phi_{\mathrm{OVAR}}$ after correction, Label Acc and Semantic Similarity of corrected text labels compared to the ground truth. \underline{PySpellChecker} is a uni-modal statistical model that makes corrections of each word based on edit distance and frequency of appearance. \underline{Bert (NeuSpell)}~\cite{Jayanthi2020NeuSpellAN} employs a uni-modal Bert-based model to translate the noisy text labels into clean ones. We ask \underline{GPT\,3.5} to give the denoised text label using the prompt ``The following words may contain spelling errors by deleting, inserting, and substituting letters. You are a corrector of spelling errors. Give only the answer without explication. What is the correct spelling of the action of <text label>?''. Our method outperforms all other methods by a considerable margin when noise rate is smaller than 20\% and is only slightly lower than GPT\,3.5 when the noise rate becomes higher. This is impressive, especially because our method is unsupervised and requires no prior knowledge. Note that GPT\,3.5, though struggles to recover the original text labels, is able to provide corrections of high semantic similarity with the ground truth, leading to one strong performance under heavy noise rate. However, the output of GPT\,3.5 tends to be unstable and relies on the design of prompts, which requires manual cleaning to remove irrelevant parts contained in the output, thus impeding real-world usage.

\vspace{0.2cm}
\noindent \textbf{Comparisons Across Datasets and Models.} 
To further determine how our method generalizes to other $\Phi_{\mathrm{OVAR}}$ models and datasets, Table~\ref{table:main} compares Top-1 Acc, showing that our model is scalable across various datasets and models. Under various noise rates, our model is robustness to guarantee small accuracy losses when applied to clean text labels and huge improvements in accuracy when applied to noisy ones. In terms of scalability across models, our method is not only applicable to hand-crafted prompts as in ActionCLIP but also to learnable visual-conditioned prompts as in XCLIP. Furthermore, we notice that, whenever XCLIP outperforms ActionCLIP, our method also yields a better result. A better visual encoder and well-tuned prompt may significantly increase the performance of our method, showing that the power of our method will increase as the community continues to train better OVAR models.

\subsection{Ablation Study}  \label{sec:ablation}

\begin{table}[t]
\centering
\caption{\textbf{Ablations for Inter-modal Weighting $\Phi_{\mathrm{Inter}}$, Intra-modal Weighting $\Phi_{\mathrm{Inter}}$, Schedule of Temperature $\lambda$}. $\Phi_{\mathrm{Inter}}$ alone outperforms $\Phi_{\mathrm{Intra}}$. Both contribute to correcting class texts, and give the best results when combined. Linear schedule of balancing factor $\lambda$ outperforms the constant one, meaning that it helps to rely more on $\Phi_{\mathrm{Intra}}$ at first, and then gradually switch to $\Phi_{\mathrm{Inter}}$.}
\vspace{-0.2cm}
\begin{tabular}{C{0.8cm}|C{0.8cm}C{0.8cm}C{1.7cm}|C{1.7cm}C{1.7cm}C{3cm}}
\toprule 
& $\Phi_{\mathrm{Inter}}$ & $\Phi_{\mathrm{Intra}}$ & $\lambda$ schedule & Top-1 Acc & Label Acc & Semantic Similarity  \\ \hline \hline
A1 &                 & $\checkmark$    & /           & $48.1_{\pm2.2} $ & $38.2_{\pm2.5} $ & $88.9_{\pm0.4} $ \\
A2 & $\checkmark$    &                 & /           & $52.9_{\pm1.4} $ & $34.1_{\pm2.4} $ & $89.1_{\pm0.6} $ \\ 
A3 & $\checkmark$    & $\checkmark$    & constant  & $54.5_{\pm2.5}  $ & $54.9_{\pm4.5}  $ & $92.4_{\pm0.8}  $ \\
A4 & $\checkmark$    & $\checkmark$    & linear      & \boldsymbol{$55.2_{\pm1.5} $} & \boldsymbol{$55.1_{\pm3.0} $} & \boldsymbol{$92.9_{\pm0.6} $} \\
\bottomrule 
\end{tabular}
\label{table:Inter_Intra}
\end{table}

\noindent {\bf Inter-modal Weighting \& Intra-modal Weighting.} 
Table \ref{table:Inter_Intra} studies how Inter-modal or Intra-modal weighting influences the performance of our model. Both contribute to denoising the text labels and to improving the robustness of underlying $\Phi_{\mathrm{OVAR}}$. We find that, in terms of Top-1 Acc and Semantic Similarity, $\Phi_{\mathrm{inter}}$ performs better than $\Phi_{\mathrm{intra}}$ as $\Phi_{\mathrm{inter}}$ optimizes in the semantic space use visual information, which is also a more direct objective of optimization to improve Top-1 Acc. On the other hand, $\Phi_{\mathrm{intra}}$ performs better in terms of Label Acc, which is more concerned with how correct the spelling is. These two modules turn out to be complementary: visual information helps the understanding of noisy text labels; while the textual information prevents the model from being misled by visual samples. Two modules perform much better when combined in terms of all of these three objectives.

\begin{figure}[t]
\centering
\vspace{0.3cm}
\includegraphics[width=0.75\linewidth]{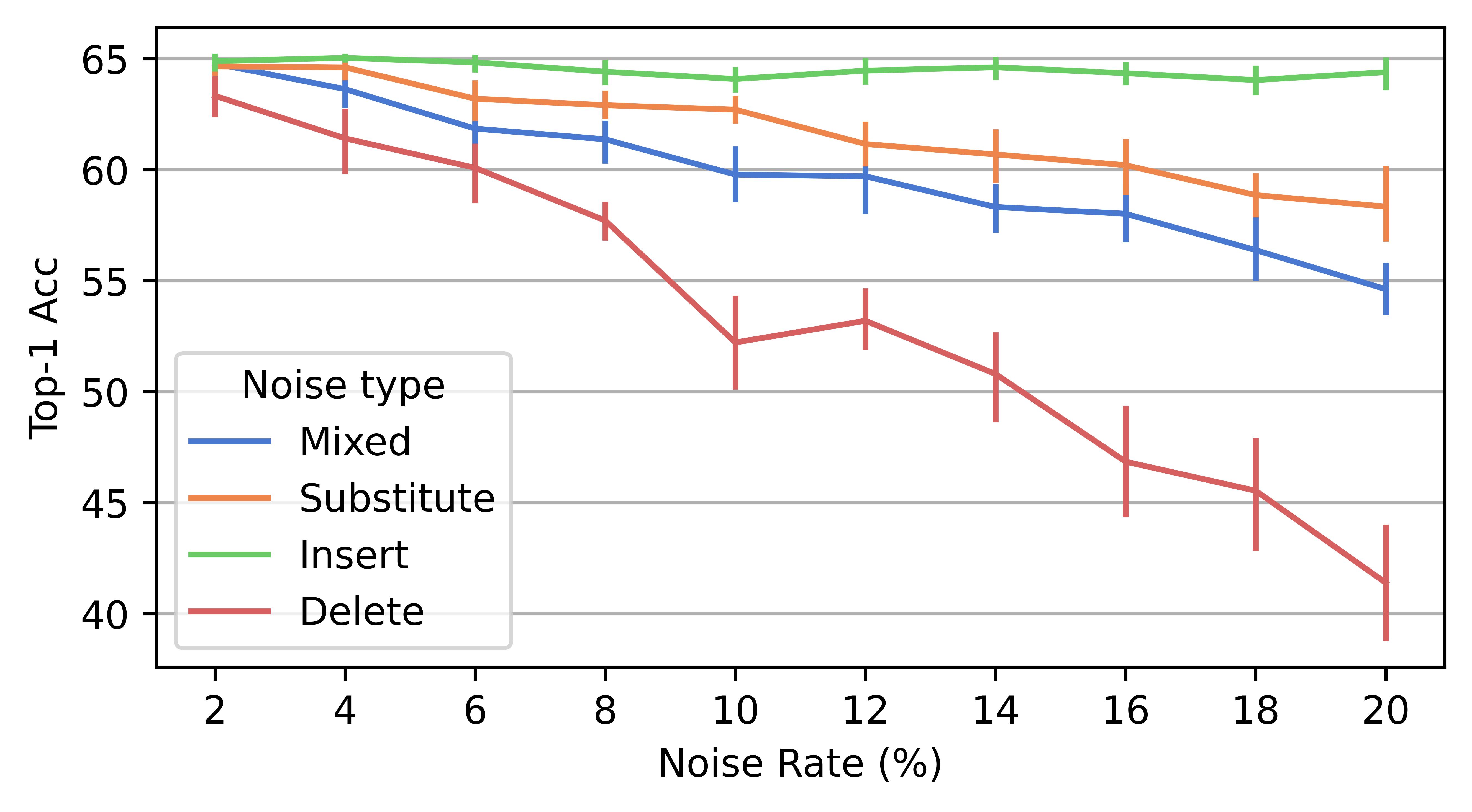}
\vspace{-0.3cm}
\caption{\textbf{Ablation Study on Noise Type}. We evaluate the robustness of our model on UCF101 with ActionCLIP as $\Phi_{\mathrm{OVAR}}$. "Mixed" means that all three types of perturbation: "Substitute", "Insert", "Delete" take place with equal probability. Our framework shows good resilience, especially against the noises of inserting or substituting.
}
\vspace{-0.3cm}
\label{fig:ablation_type_nosie}
\end{figure}

\vspace{0.2cm}
\noindent {\bf Type of Noise.}
Fig~\ref{fig:ablation_type_nosie} studies the robustness under different noise type and noise rate. “Mixed” means that the three noise types: ``Substitute'', ``Insert'', ``Delete'' are equally possible to appear. Our method shows remarkable resilience when the text labels are perturbed by inserting or substituting characters. A decrease in performance is observed when text labels are perturbed by deleting characters. It is understandable that by deleting characters, we suffer from huge information loss, making the model difficult to recover the original text labels.

\vspace{0.2cm}
\noindent {\bf Schedule of Temperature $\lambda$.}
Temperature schedule $\lambda$ adjusts the importance of Intra-modal weighting over Inter-modal weighting. One larger $\lambda$ indicates that we rely more on Inter-modal weighting, and vice versa. 
``linear'' means that $\lambda$ augments from 0.01 to 1 linearly. Table \ref{table:Inter_Intra} reports that it is beneficial to rely more on Intra-modal information at the beginning of the decoding process, and then gradually turn to Inter-modal information for more help. We hypothesize that with the performance of the OVAR model $\Phi_{\mathrm{OVAR}}$ limited due to noisy text labels, $\Phi_{\mathrm{inter}}$ offers more help than $\Phi_{\mathrm{intra}}$; and that with $\Phi_{\mathrm{OVAR}}$ performances better and better, $\Phi_{\mathrm{inter}}$ could outperform $\Phi_{\mathrm{intra}}$.

\begin{figure}[t]
\centering
\includegraphics[width=0.75\linewidth]{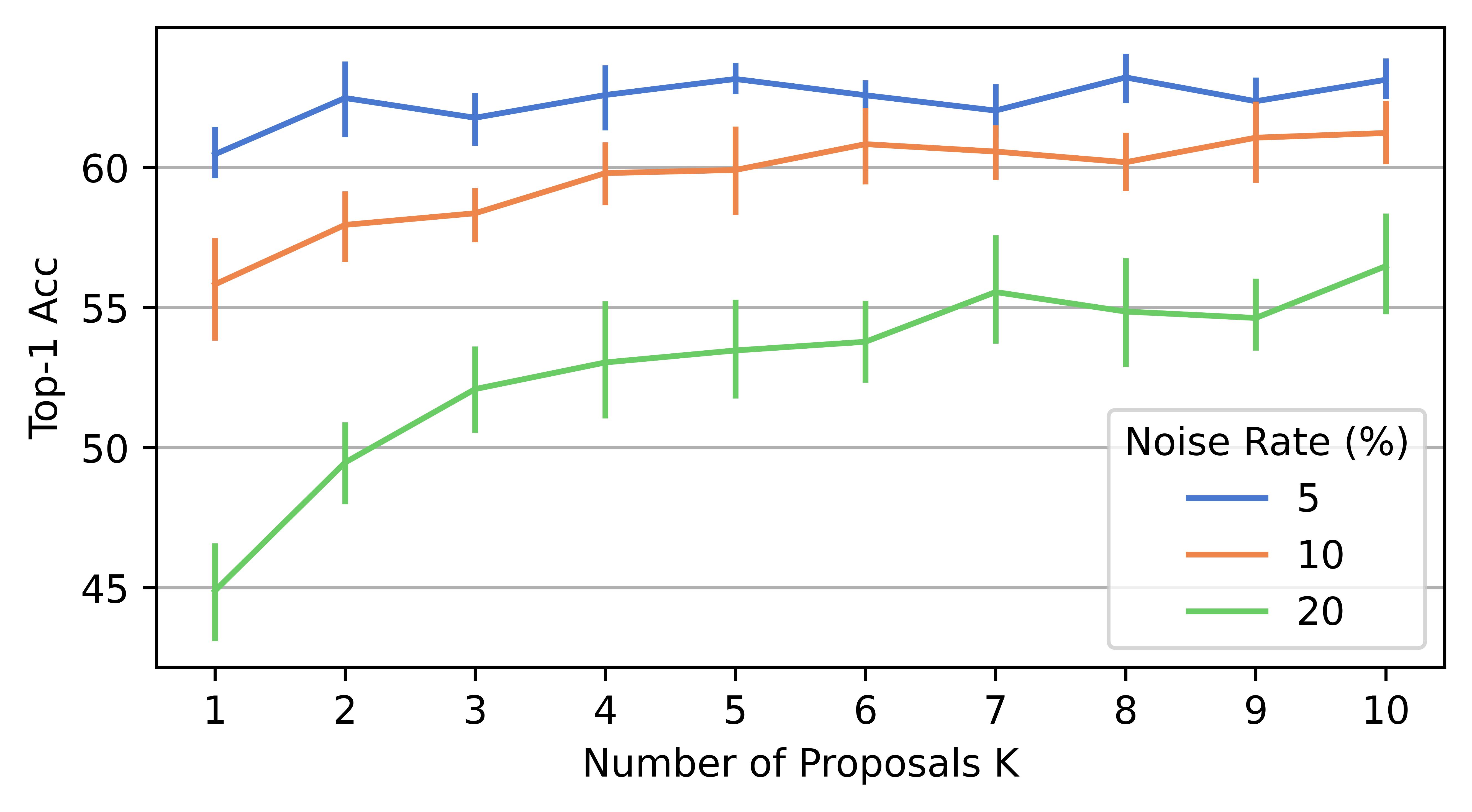}
\vspace{-0.3cm}
\caption{\textbf{Ablation Study for Proposal Number} $\mathbf{K}$. We evaluate on UCF101 using ActionCLIP as $\Phi_{\mathrm{OVAR}}$. As proposal number $K$ increases, the Top-1 Acc increases and converges gradually towards the upper bound, but also brings heavier computing costs.}
\vspace{0.4cm}
\label{fig:ablation_topk}
\end{figure}

\vspace{0.15cm}
\noindent {\bf Number of Candidates.} 
Fig.~\ref{fig:ablation_topk} shows as the number of proposed candidates $K$ increases, the Inter-modal weighting may reveal its full power, hence increasing performance. Otherwise, if a good candidate is excluded in the proposal stage because of a small $K$, it can be selected by neither of the inter or intra modal weighting, hence decreasing performance. However, the performance tends towards a plateau, showing a decreasing marginal contribution of more proposals to the performance. Since a larger $K$ means more computational cost while encoding the text, we select $K=10$ by default to make a reasonable compromise.

\begin{table}[t]
\centering
\caption{\textbf{Cases of Denoised Text Labels between GPT\,3.5 and \textit{DENOISER}}. The output of GPT\,3.5~\cite{gpt3.5} tends to be unstable and represents sometimes a relatively high-level understanding of noisy class-text labels. Our \textit{DENOISER} framework ensures a relatively faithful output in terms of spelling but could be slightly mistaken when two words are similar in terms of both semantics and spelling.}
\vspace{-0.3cm}
\begin{tabular}{C{1cm}|C{3cm}C{3cm}|C{2cm}C{3cm}}
\toprule
                           & Ground Truth & Noise Class Labels & GPT\,3.5~\cite{gpt3.5} & \textbf{Ours} \\ \hline  \hline
\multirow{3}{*}{\shortstack[c]{Good \\ Case}}
                           & walking with a dog & wal\textcolor{red}{4}ing\textcolor{red}{m} with a dog & dogwalking & \textcolor{green}{walking} with a dog\\
                           & baby crawling & bab\textcolor{red}{t}y crawling & baby crying & \textcolor{green}{baby} crawling\\

                           & cutting in kitchen & cutting i\textcolor{red}{\underline{\hspace{0.2cm}}} \textcolor{red}{a}it\textcolor{red}{n}chen & cutting & cutting \textcolor{green}{in} \textcolor{green}{kitchen}\\
                           \midrule
\shortstack[c]{Bad \\ Case}
& juggling balls & juggling ball\textcolor{red}{\underline{\hspace{0.2cm}}} & juggling & juggling \textcolor{green}{ball}\textcolor{red}{\underline{\hspace{0.2cm}}} \\
\bottomrule
\end{tabular}
\label{table:goodbad}
\end{table}

\begin{figure}[t]
    \centering
    \begin{minipage}{0.32\linewidth}
        \centering
        \includegraphics[width=0.99\linewidth]{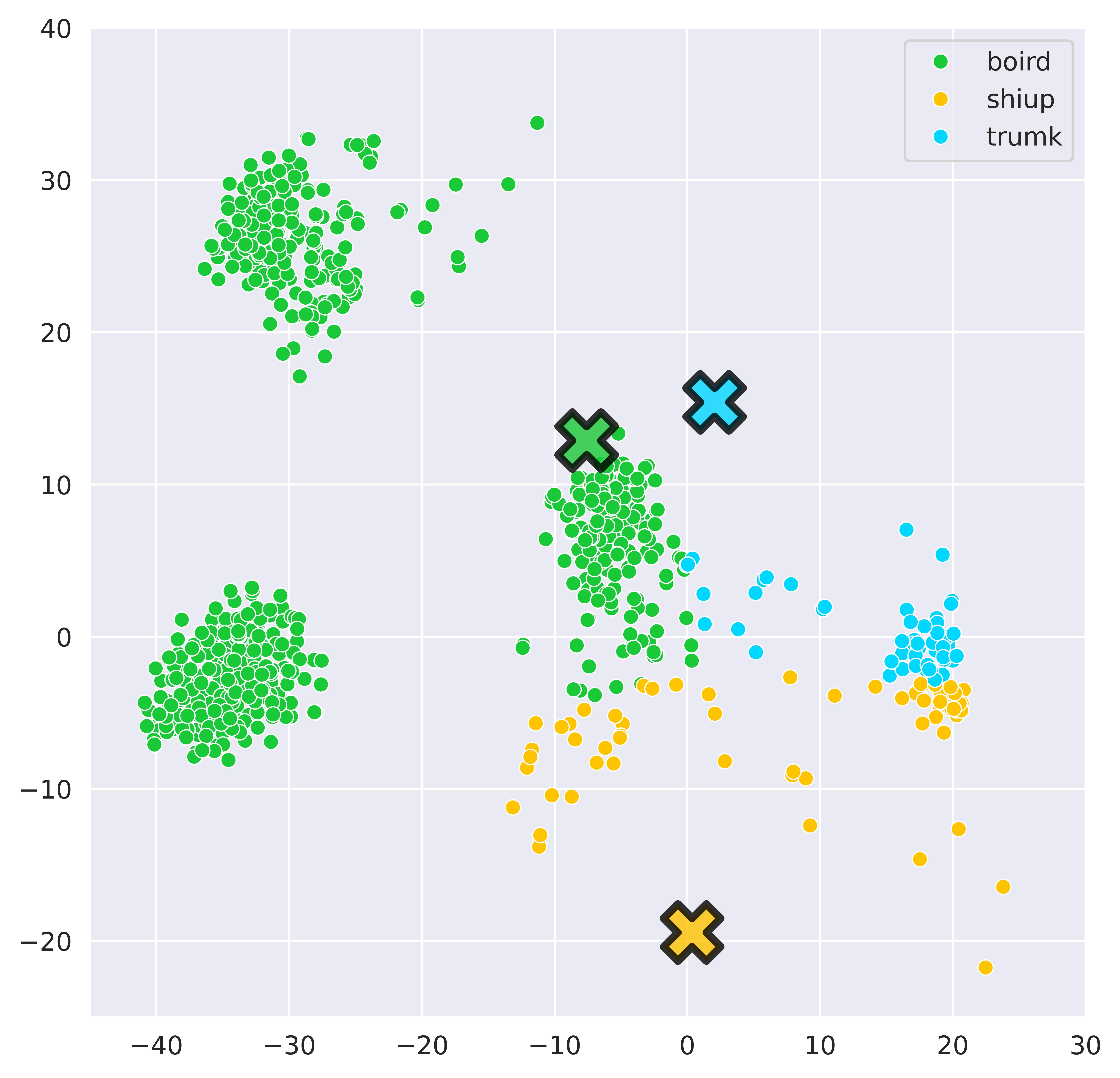}
        \label{noisy1}
    \end{minipage}
    \begin{minipage}{0.32\linewidth}
        \centering
        \includegraphics[width=0.99\linewidth]{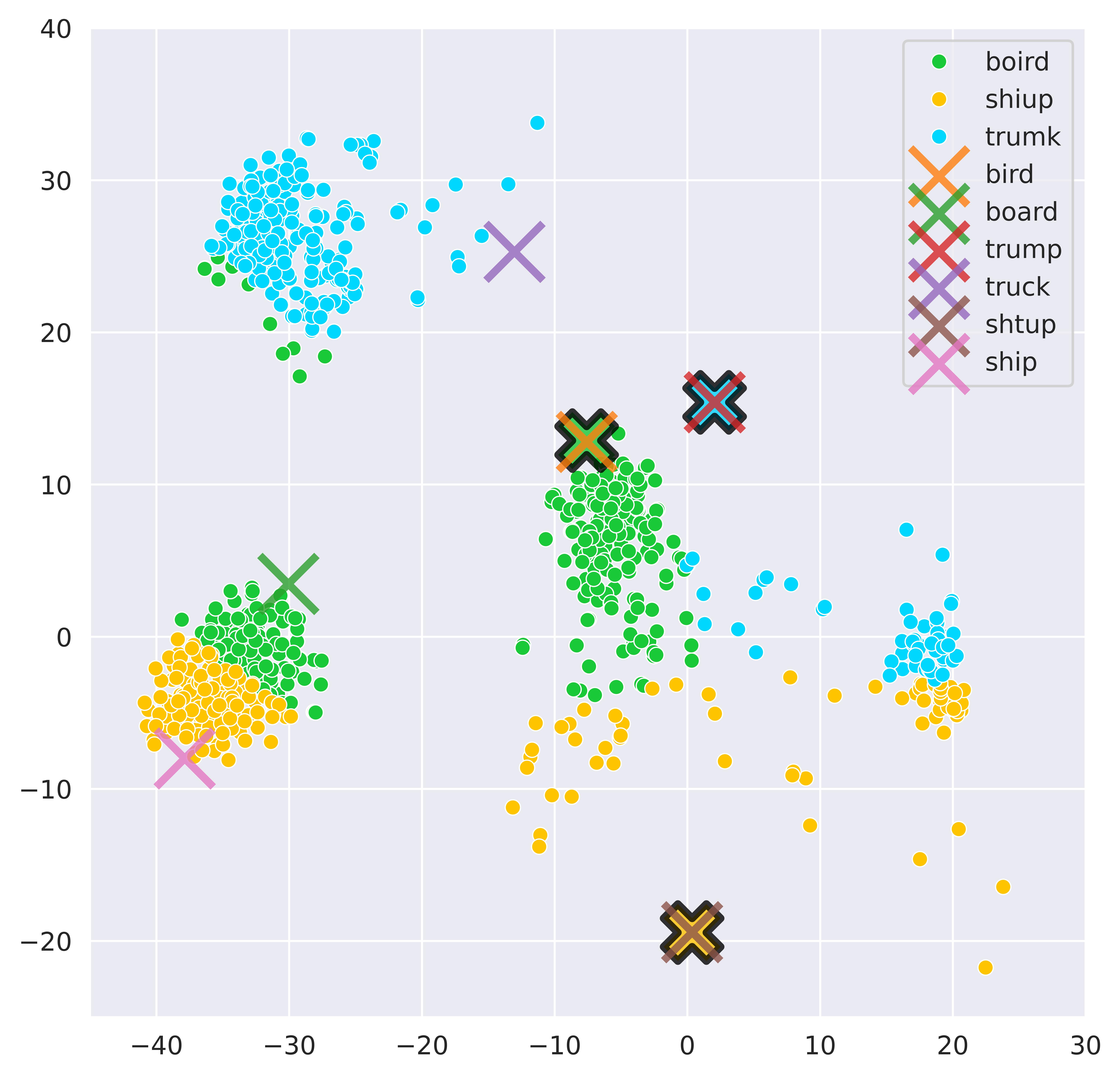}
        \label{noisy2}
    \end{minipage}
    \begin{minipage}{0.32\linewidth}
        \centering
        \includegraphics[width=0.99\linewidth]{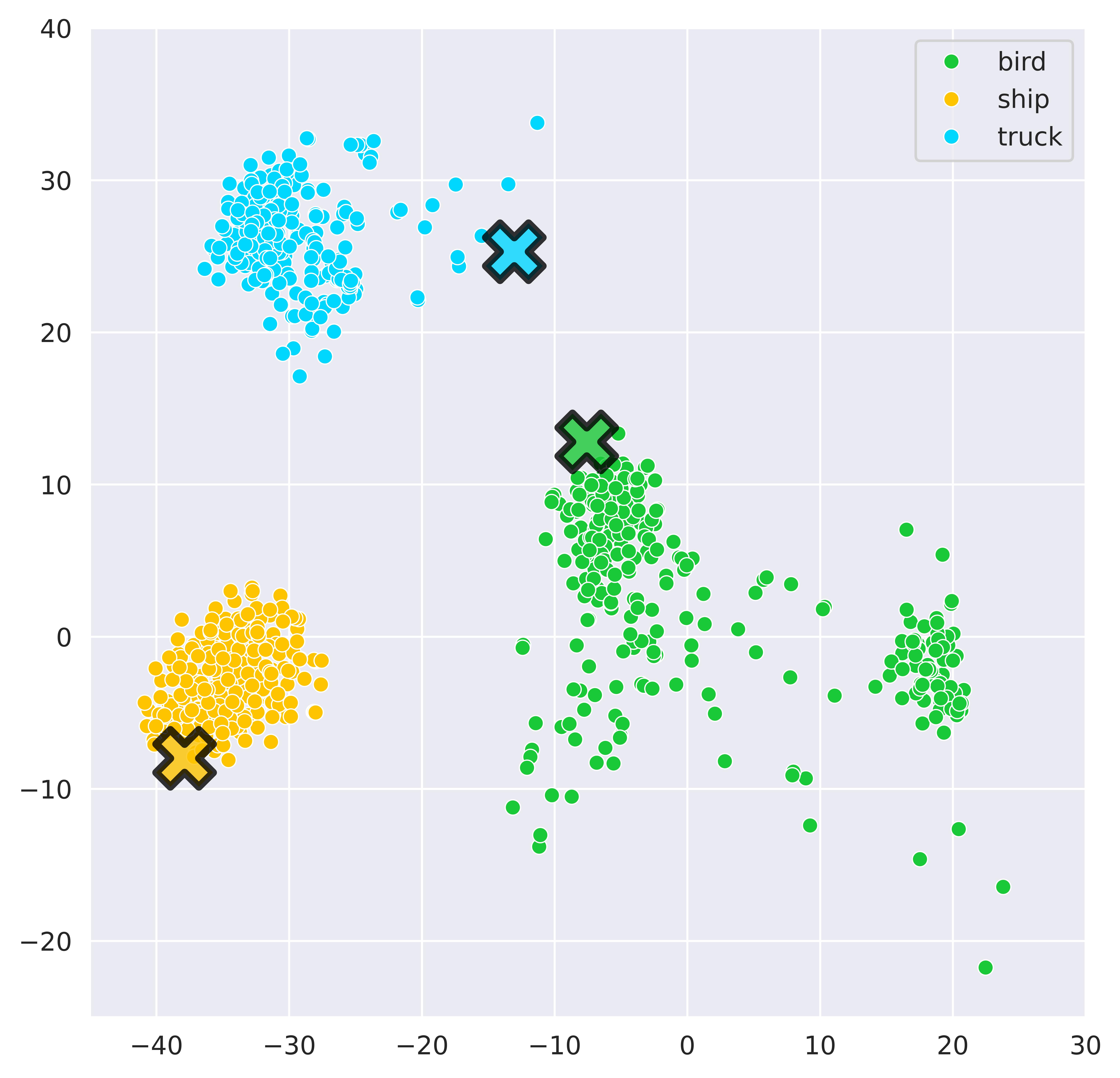}
        \label{noisy3}
    \end{minipage}
    \vspace{-0.5cm}
  \caption{\textbf{Visualization of Denoising Process.} \textbf{Left:} classification result with noisy text labels (in crosses with black border). \textbf{Middle:} text candidates (in crosses without black border), the visual samples (in dots) that are used to vote for candidates. \textbf{Right:} denoised class texts (in crosses with black border) help for better classification.}
  \label{fig:noisy}
\end{figure}

\subsection{Qualitative Results}  \label{sec:quality}
In Fig.~\ref{fig:noisy}, we visualize the embedding of visual samples and corresponding text labels of three categories: bird (green), ship (yellow), truck (blue) in CIFAR-10 using T-SNE. The first principal component of textual embedding is removed following ReCLIP\cite{hu2024reclip} to prevent them from clustering at the same place. Fig.~\ref{fig:noisy} Left shows that classification accuracy is low when the text labels are noisy. Almost all visual samples are classified to be ``bird''. Fig.~\ref{fig:noisy} Middle shows the embeddings of proposed text candidates. Note that some of them remain at the same place. We hypothesize that they move perpendicular to this 2D space in the real semantic space. We assign the best set of visual samples for each category to help denoise, {\em e.g.}, the blue dots are used to vote on the two candidates ``trump'' (red) and ``truck'' (purple) of ``trumk''. Fig.~\ref{fig:noisy} Right shows that the denoised text labels improve the performance of OVAR model.

In Table~\ref{table:goodbad}, we qualitatively compare several of the good/bad cases for our method and GPT\,3.5. We find that GPT\,3.5 is better at understanding the semantics of noisy class labels, {\em e.g.}, ``wal4ingm with a dog'' $\rightarrow$ ``dogwalking''. However, its output tends to be unstable, omitting or misunderstanding sometimes an important part of a class label, {\em e.g.}, ``babty crawling'' $\rightarrow$ ``baby crying''. Note that the output of GPT\,3.5 is greatly influenced by the prompt and often requires manual cleaning due to unstable output, limiting its capacity in real-world applications. Our \textit{DENOISER} framework remains more faithful in terms of spelling, {\em e.g.}, ``wal4ingm with a dog'' $\rightarrow$ ``walking with a dog'' instead of $\rightarrow$ ``dogwalking''. However, it could be mistaken when two words are similar in terms of both semantics and spelling, such as ``ball'' and ``balls''.

\section{Conclusion}
This paper first investigates how noises in class texts negatively affect the performance for Open-Vocabulary Action Recognition, {\em i.e.}, OVAR. We propose \textit{DENOISER}, a novel framework for solutions. By incorporating visual information during denoising, we clarify the ambiguity induced by short and context-lacking class texts; by iteratively refining the denoised output through one generative-discriminative process, we mitigate cascaded errors which may propagate from spell-checking models to outputs of OVAR model. We conduct extensive experiments to demonstrate the generalizability of \textit{DENOISER} across existing models and datasets as well as its superiority over uni-modal spell-checking models.

\section{Appendix}
\subsection{Decoding Objective}
At each step $i$, the decoding objective to find $\argmax_{t_i} p(t_{i} |\mathcal{T}_{i-1}, \mathcal{V}) $. Note that, $p(\mathcal{T}_{i-1}, \mathcal{V})$ is same for all possible $t_i$. As a result, our objectify is written as:
\begin{align}
\argmax_{t_i} p(t_{i} |\mathcal{T}_{i-1}, \mathcal{V}) &= \argmax_{t_i} p(t_{i} |\mathcal{T}_{i-1}, \mathcal{V})p(\mathcal{T}_{i-1}, \mathcal{V}) \\
&= \argmax_{t_i} p(t_{i} ,  \mathcal{T}_{i-1}, \mathcal{V}) \\
&= \argmax_{t_i} \log p(t_{i} ,  \mathcal{T}_{i-1}, \mathcal{V})
\end{align}

\subsection {Discriminative Step} 
At discriminative step, we choose the best set of $\mathcal{V}$ that helps decode $t_{c,i}$ for each category $c$. To understand why $\mathcal{V}_c$, the set of visual samples $v_j$ whose labels $\mathcal{Y}_j$ are assigned to category $c$ are those who help decode most efficiently, we first introduce a hidden discrete random variable $z_j \sim Q_j$ for each $v_j$, indicating the index of class assignment. $z_j = c$ means that $\argmax \mathcal{Y}_j = c$.

Knowing that all visual samples are independent and using Jensen inequality:
\begin{align}
\log p(t_i, \mathcal{T}_{i-1}, \mathcal{V}) & = \sum_j \log p(t_i, \mathcal{T}_{i-1}, v_j) \\
& = \sum_j \log \sum_{z_j} p(t_i, \mathcal{T}_{i-1}, v_j, z_j) \\
& = \sum_j \log \sum_{z_j} Q_{j}(z_j) \frac{p(t_i, \mathcal{T}_{i-1}, v_j, z_j)}{Q_{j}(z_j)} \\
& \geq \sum_j \sum_{z_j} Q_{j}(z_j) \log \frac{p(t_i, \mathcal{T}_{i-1}, v_j, z_j)}{Q_{j}(z_j)}
\end{align}

Equality is attained at $Q_{j}(z_j) \varpropto p(t_i, \mathcal{T}_{i-1}, v_j, z_j)$. Since $\sum_{z_j} Q_j(z_j) = 1$, to maximize the lower bound, we have:
\begin{align}
Q_j(z_j) & = \frac{p(t_i, \mathcal{T}_{i-1}, v_j, z_j)}{\sum_{z_j} p(t_i, \mathcal{T}_{i-1}, v_j, z_j)} \\
& = \frac{p(t_i, \mathcal{T}_{i-1}, v_j, z_j)}{p(t_i, \mathcal{T}_{i-1}, v_j)} \\
& = p(z_j | t_i, \mathcal{T}_{i-1}, v_j) \\
& = p(z_j | \mathcal{T}_{i}, v_j)
\end{align}

Given class texts and visual samples, the best estimation is:
\begin{align}
\mathbb{P}(z_j = c|\mathcal{T}_i,v_j) =
    \begin{cases}
        1 \qquad & c = \argmax\limits_c \max\limits_k \frac{\text{exp}(\mathcal{S}(v_j,\mathcal{T}^k_{c, i}))}{\sum_{k^\prime} \text{exp}(\mathcal{S}( v_j,\mathcal{T}^{k^\prime}_{c, i}))} \\
        0 \qquad & \text{otherwise} \\
    \end{cases}
\end{align}

Note that, $Q_j$ is well defined because:
\begin{align}
\lim_{Q_{j}(z_j)\to 0^+} Q_{j}(z_j) \log \frac{p(t_i, \mathcal{T}_{i-1}, v_j, z_j)}{Q_{j}(z_j)} = 0
\end{align}

With $Q_j$ defined in this way, we find the discriminative step to be identical to how $\Phi_{\mathrm{OVAR}}$ assigns labels. We have $Q_{j}(c) = 1$ only for $\{j|v_j\in \mathcal{V}_c\}$:
\begin{align}
\log p(t_i, \mathcal{T}_{i-1}, \mathcal{V}) & \geq \sum_j \sum_{z_j} Q_{j}(z_j) \log \frac{p(t_i, \mathcal{T}_{i-1}, v_j, z_j)}{Q_{j}(z_j)} \\
& = \sum_c \sum_{j, v_j\in \mathcal{V}_c} \sum_{z_j} Q_{j}(z_j) \log \frac{p(t_i, \mathcal{T}_{i-1}, v_j, z_j)}{Q_{j}(z_j)} \\
& = \sum_c  \sum_{j, v_j\in \mathcal{V}_c} \log p(t_i, \mathcal{T}_{i-1}, v_j, z_j = c) \\
& = \sum_c  \log p(t_{c,i}, \mathcal{T}_{c,i-1}, \mathcal{V}_c) \\
\end{align}

\subsection {Generative Step}
We optimize $t_{c,i}$ for each category:
\begin{align}
\argmax_{t_{c,i}} \log p(t_{c,i} ,  \mathcal{T}_{c,i-1}, \mathcal{V}_c) & = \argmax_{t_{c,i}} p(t_{c,i}, \mathcal{T}_{c,i-1}, \mathcal{V}_c) \\
& = \argmax_{t_{c,i}} \prod_{v_j\in\mathcal{V}_c} p(t_{c,i}, \mathcal{T}_{c,i-1}, v_j) \\
&= \argmax_{t_{c,i}} \prod_{v_j\in\mathcal{V}_c} p(\mathcal{T}_{c, i-1}|t_{c,i}, v_j) p(t_{c,i}|v_j) p(v_j) \\
&= \argmax_{t_{c,i}} \prod_{v_j\in\mathcal{V}_c} p(\mathcal{T}_{c, i-1}|t_{c,i}, v_j) p(t_{c,i}|v_j)
\end{align}

Noting that $p(\mathcal{T}_{c, i-1})$ is the same for any possible $t_{c,i}$:
\begin{align}
\argmax_{t_{c,i}} p(\mathcal{T}_{c, i-1}|t_{c,i}, v_j) &= \argmax_{t_{c,i}} p(\mathcal{T}_{c, i-1}|t_{c,i}) \\
&= \argmax_{t_{c,i}}\frac{p(t_{c,i}|\mathcal{T}_{c, i-1})p(\mathcal{T}_{c, i-1})}{p(t_{c,i})}\\
&= \argmax_{t_{c,i}} \frac{p(t_{c,i}|\mathcal{T}_{c, i-1})}{p(t_{c,i})} 
\end{align}

It is possible to optimize with prior $p(t_{c,i})$ by considering that the more a word is frequent, the less it is likely to be misspelled in real-world scenarios. In this paper, for simplicity, we assume the $t_{c,i}$ to be uniform:
\begin{align}
\argmax_{t_{c,i}} p(\mathcal{T}_{c, i-1}|t_{c,i}, v_j) &= \argmax_{t_{c,i}} p(t_{c,i}|\mathcal{T}_{c, i-1})
\end{align}

\subsection{DENOISER \textit{vs}. Adversarial Training}  \label{sec:adv train} 
\begin{figure}[h]
\centering
\includegraphics[width=\linewidth]{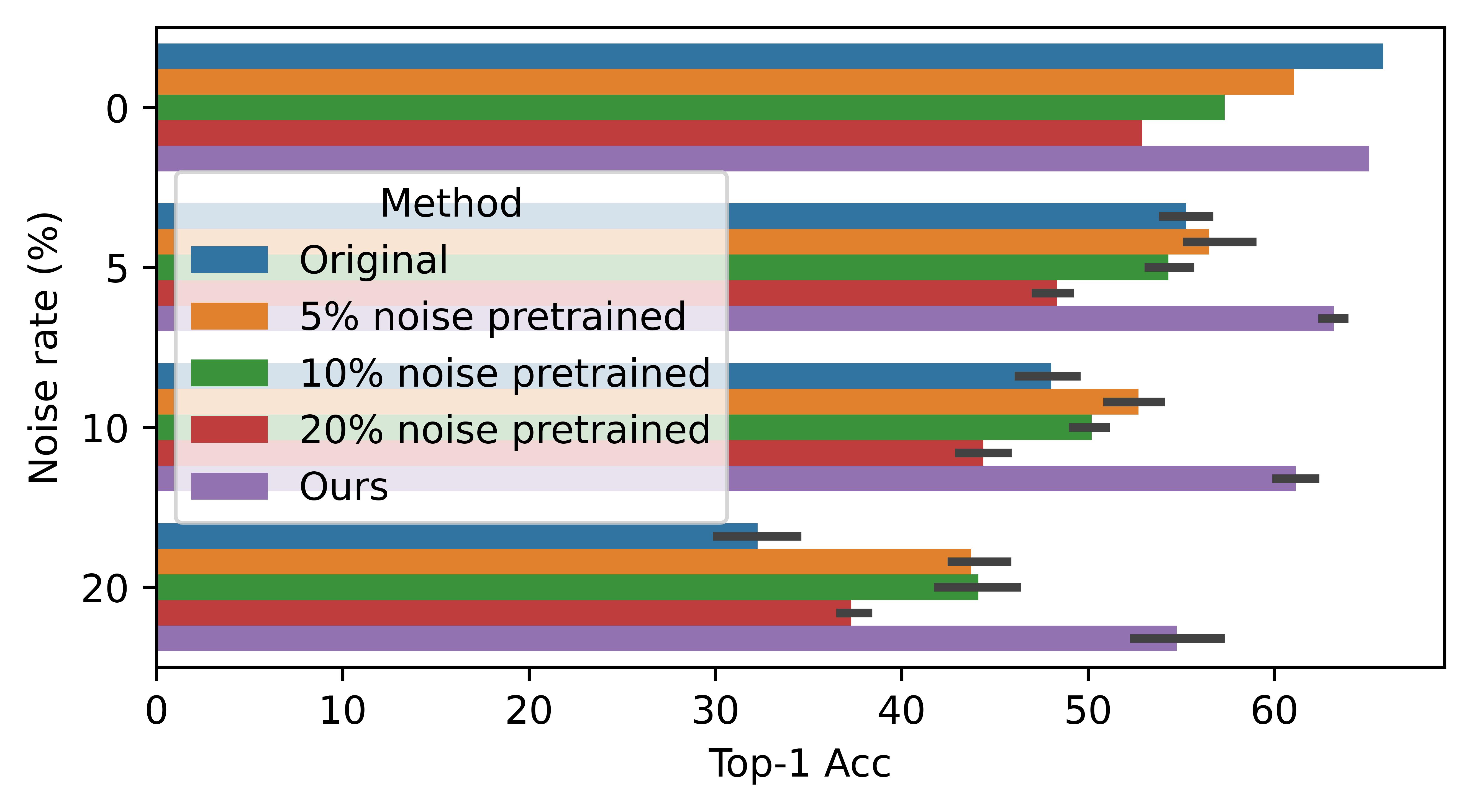}
\caption{\textbf{Comparison to Adversarial Training.} Adversarial training is not efficient, especially in low-noise scenarios, even leading to a lower performance compared to the original model. It also falls behind our method by a significant margin.}
\label{fig:adversarial}
\end{figure}

Fig~\ref{fig:adversarial} studies how adversarial training might mitigate the noise in text labels. We first train ActionCLIP ViT-B/32-8F from scratch on K400 by randomly injecting noise in its text labels, then test the model's zero-shot performance on UCF101 under different noise rate scenarios. We find that adversarial training, though promising under closed-set scenarios in previous studies, is relatively ineffective under open-vocabulary settings. Specifically, training with more noise lowers significantly the model's performance under low noise rate. Additionally, its added value is limited under heavy noise rate. These phenomena are probably related to the domain gap between datasets. By training on noisy text labels, the model tends to overfit the noise pattern, jeopardizing its zero-shot performance. We conclude that noisy text labels are better solved in testing time rather than during training stage. Our DENOISER framework shows a significant advantage over adversarial training.

\subsection{Ablation on Number of Visual Samples Used}
\begin{figure}[h]
    \centering
    \includegraphics[width=\linewidth]{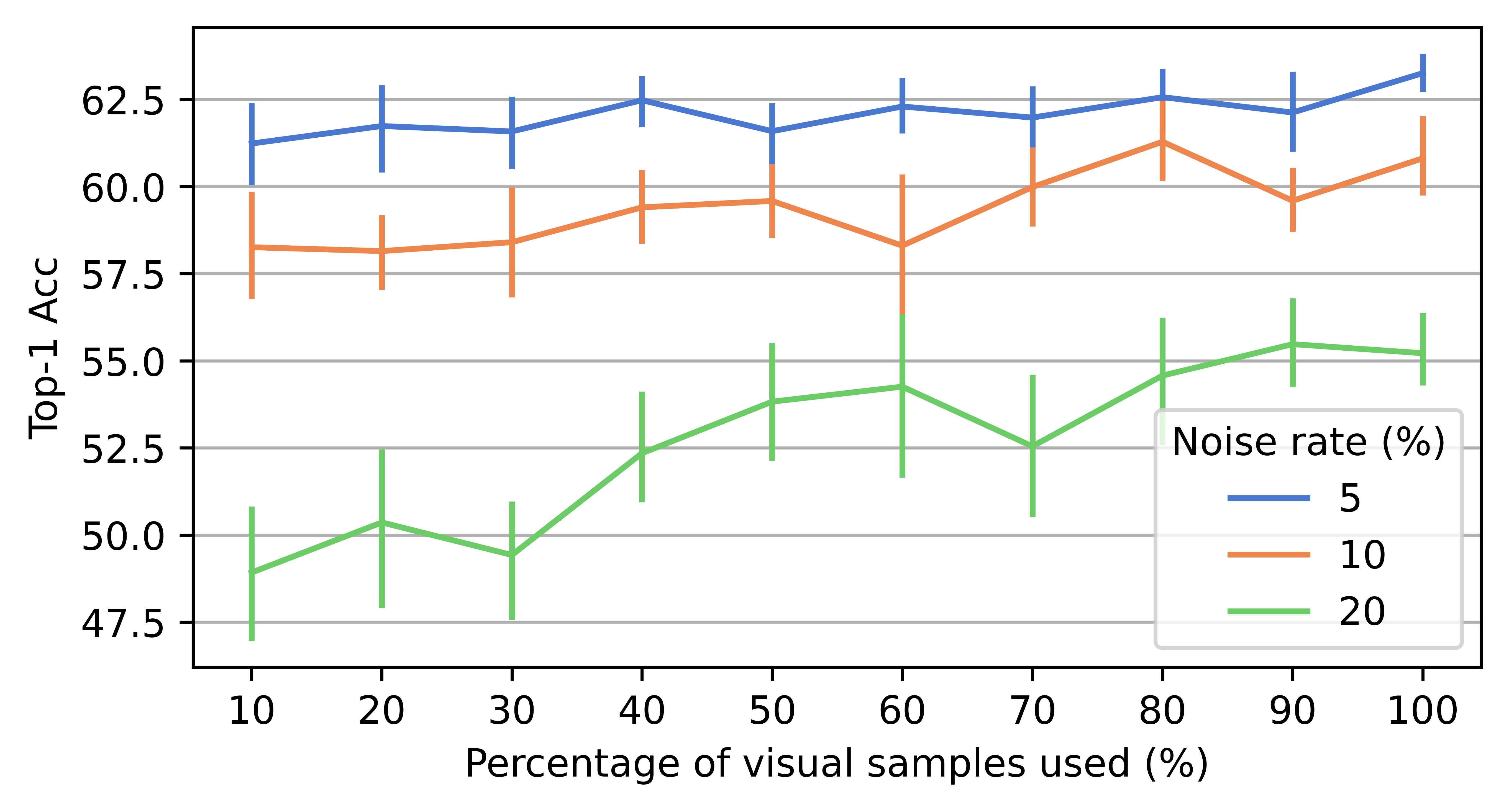}
    \caption{\textbf{Ablation on Number of Visual Samples Used}. When fewer visual samples are used in $\Phi{\mathrm{inter}}$, our method shows a drop in performance. The bigger the noise rate, the larger the drop, showing that $\Phi{\mathrm{inter}}$ plays a role of increasing importance when the noise is larger.}
    \label{fig:ablation_percentage_False}
\end{figure}

We ablate in Fig.~\ref{fig:ablation_percentage_False}. on the number of visual samples used in $\Phi{\mathrm{inter}}$. Our method shows a drop in performance when fewer visual samples are used in $\Phi{\mathrm{inter}}$. The performance tends to converge towards that when solely $\Phi{\mathrm{intra}}$ is used. We hypothesize that fewer visual samples make $\Phi{\mathrm{inter}}$ harder to extract added value to $\Phi{\mathrm{intra}}$. With the noise rate increasing, we find an increasingly large drop in performance, which shows conversely that $\Phi{\mathrm{inter}}$ is more important under large noise scenarios as textual information becomes more ambiguous and less informative.

\clearpage
\bibliographystyle{splncs04}
\bibliography{egbib}
\end{document}